\documentclass[conference]{IEEEtran}
\IEEEoverridecommandlockouts

\usepackage{cite}
\usepackage{amsmath,amssymb,amsfonts}
\usepackage{algorithmic}
\usepackage{graphicx}
\usepackage[inline]{enumitem}
\usepackage{textcomp}
\usepackage{xcolor}
\usepackage{widetable}
\usepackage{hyperref}
\usepackage{comment}
\usepackage{multirow}
\usepackage{cleveref}
\def\BibTeX{{\rm B\kern-.05em{\sc i\kern-.025em b}\kern-.08em
    T\kern-.1667em\lower.7ex\hbox{E}\kern-.125emX}}
\begin{document}

\title{Quantum Machine Learning in Healthcare: Evaluating QNN and QSVM Models 
\vspace{-10pt}
}

\author{\IEEEauthorblockN{Antonio Tudisco\IEEEauthorrefmark{1},
Deborah Volpe\IEEEauthorrefmark{1}, and  Giovanna Turvani\IEEEauthorrefmark{1} }
\IEEEauthorblockA{\IEEEauthorrefmark{1}Department of Electronics and Telecommunications,
Politecnico di Torino
Italy\\
\href{mailto:antonio.tudisco@polito.it}{antonio.tudisco@polito.it},
 \href{mailto:deborah.volpe@polito.it}{deborah.volpe@polito.it}, and
\href{mailto:giovanna.turvani@polito.it}{giovanna.turvani@polito.it}}
\vspace{-30pt}}

\maketitle
\begin{abstract} 
Effective and accurate diagnosis of diseases such as cancer, diabetes, and heart failure is crucial for timely medical intervention and improving patient survival rates. Machine learning has revolutionized diagnostic methods in recent years by developing classification models that detect diseases based on selected features. However, these classification tasks are often highly imbalanced, limiting the performance of classical models.\\
Quantum models offer a promising alternative, exploiting their ability to express complex patterns by operating in a higher-dimensional computational space through superposition and entanglement. These unique properties make quantum models potentially more effective in addressing the challenges of imbalanced datasets.
This work evaluates the potential of quantum classifiers in healthcare, focusing on Quantum Neural Networks (QNNs) and Quantum Support Vector Machines (QSVMs), comparing them with popular classical models. The study is based on three well-known healthcare datasets --- Prostate Cancer, Heart Failure, and Diabetes. \\
The results indicate that QSVMs outperform QNNs across all datasets due to their susceptibility to overfitting. Furthermore, quantum models prove the ability to overcome classical models in scenarios with high dataset imbalance. Although preliminary, these findings highlight the potential of quantum models in healthcare classification tasks and lead the way for further research in this domain.
\end{abstract}


\section{Introduction}

Effective and accurate diagnosis of diseases such as cancer, diabetes, and heart failure is crucial for timely medical intervention and improving patient survival rates. In recent years, \textbf{Machine Learning} (\textbf{ML}) has revolutionized diagnostic methods by developing classification models \cite{an2023comprehensive} that detect diseases based on selected features derived from patient data, such as biomarkers, genetic profiles, and medical imaging. These methods exploit algorithms ranging from logistic regression and support vector machines, enabling automated systems to identify patterns that may be difficult for human experts to distinguish.
However, these classification tasks are often highly imbalanced, as the prevalence of the disease is much lower than that of healthy cases or vice versa in the datasets used for training. This imbalance can lead to biased models that favor the majority class, resulting in poor sensitivity and low recall for the minority (disease) class. Consequently, addressing data imbalance has become a key point in improving model reliability \cite{haixiang2017learning}.

Quantum models offer a promising alternative to classical ML techniques \cite{10398184}, exploiting their ability to express complex and intricate patterns by operating in a higher-dimensional computational space obtained through the principles of superposition and entanglement. These unique quantum properties allow for enhanced feature representation and the ability to explore complex data, potentially making quantum models more effective in addressing challenges such as imbalanced datasets \cite{tudiscoevaluating2024}, which is a common limitation in healthcare applications.

This work evaluates the potential of quantum classifiers in healthcare diagnostics, focusing specifically on \textbf{Quantum Neural Network} (\textbf{QNNs}) \cite{Abbas2021-yv} and \textbf{Quantum Support Vector Machines} (\textbf{QSVMs})\cite{schuld2021supervised} --- two widely studied Quantum Machine Learning (QML) approaches. QNNs utilize a \textbf{Variational Quantum Circuit} (\textbf{VQC}) \cite{schuld2020circuit}, i.e. a parameterized quantum circuit optimized through iterative training, providing a flexible and expressive framework for learning complex patterns. QSVMs, on the other hand, leverage on quantum kernel methods to project data into higher-dimensional feature spaces, enabling more effective classification with fewer computational resources.

Moreover, we compared the quantum models with their classical counterparts, considering Logistic Regression, Decision Tree, Random Forest, and Support Vector Machine (SVM).
The evaluation was conducted using three well-known healthcare datasets --- Prostate Cancer \cite{prostate_cancer}, Heart Failure \cite{heart_failure_clinical_records_519}, and Diabetes \cite{diabetes_dataset} --- representing different diagnostic challenges. To address the limitations posed by current Noisy Intermediate-Scale Quantum (NISQ) devices and quantum emulators, a \textbf{Principal Component Analysis} (\textbf{PCA}) \cite{PCA} preprocessing step was applied to reduce the dimensionality of the datasets. This step ensured compatibility with quantum hardware constraints while preserving the most informative features, thus enhancing the performance and scalability of the quantum classifiers.

The experimental results indicate that QSVMs outperform QNNs across all datasets. While QNNs demonstrated the ability to model complex patterns, they were affected by overfitting, probably due to the need for careful hyperparameter tuning. In contrast, QSVMs benefited from their inherent ability to handle nonlinear decision boundaries and class imbalance, producing more stable and generalizable performance. Furthermore, quantum models outperform classical models on the dataset with the highest imbalance, whereas classical models perform better on the most balanced dataset. This demonstrates that \textbf{quantum models become increasingly advantageous as dataset imbalance increases}. 
Although preliminary, these results highlight the potential of quantum models in healthcare classification tasks, particularly for datasets with imbalanced distributions and intricate feature interactions. They underscore the importance of further research into hybrid quantum-classical approaches and the exploration of more advanced quantum kernels. Such advancements could lead to the exploitation of quantum algorithms in real-world medical applications, improving diagnostic quality.

The rest of the article is organized as follows. 
Section~\ref{sec:MotivationGen} outlines the motivation behind this work and the general idea, while Section~\ref{sec:Implementation} presents the actual models implementation and their characteristics. The obtained results are presented and discussed in Section~\ref{sec:Results}. Finally, in Section~\ref{sec:conclusions}, conclusions are drawn, and future perspectives are illustrated.
 \section{Motivation and General Idea} \label{sec:MotivationGen}
 Healthcare diagnostic classification tasks are usually characterized by \textbf{imbalanced datasets} --- positive cases (e.g., diseased patients) occur less frequently than negative ones (e.g., healthy individuals) ---, representing a challenge for ML models. Quantum models have the potential to overcome the limits of their classical counterparts in case of imbalanced classification since they operate in \textbf{higher-dimensional Hilbert space} thanks to superposition and entanglement. 

This work explores the potential of quantum classifiers in healthcare, considering two popular near-term learning paradigms, i.e., \textbf{QNN} and \textbf{QSVM}, and the impact of their \textbf{hyperparameters} on disease detection quality. Logistic Regression, Decision Tree, Random Forest, and Support Vector Machine (SVM) were selected for comparison with classical models. Three different datasets --- Prostate Cancer, Heart Failure, and Diabetes --- were chosen for their heterogeneity, as they exhibit differences in class imbalance, number of samples, and feature characteristics. This diversity allows for the evaluation of quantum classifiers in varied scenarios and supports drawing more general conclusions about their effectiveness in healthcare applications.
\begin{figure}[h]
    \centering \vspace{-15pt}
    \includegraphics[width=0.8\linewidth]{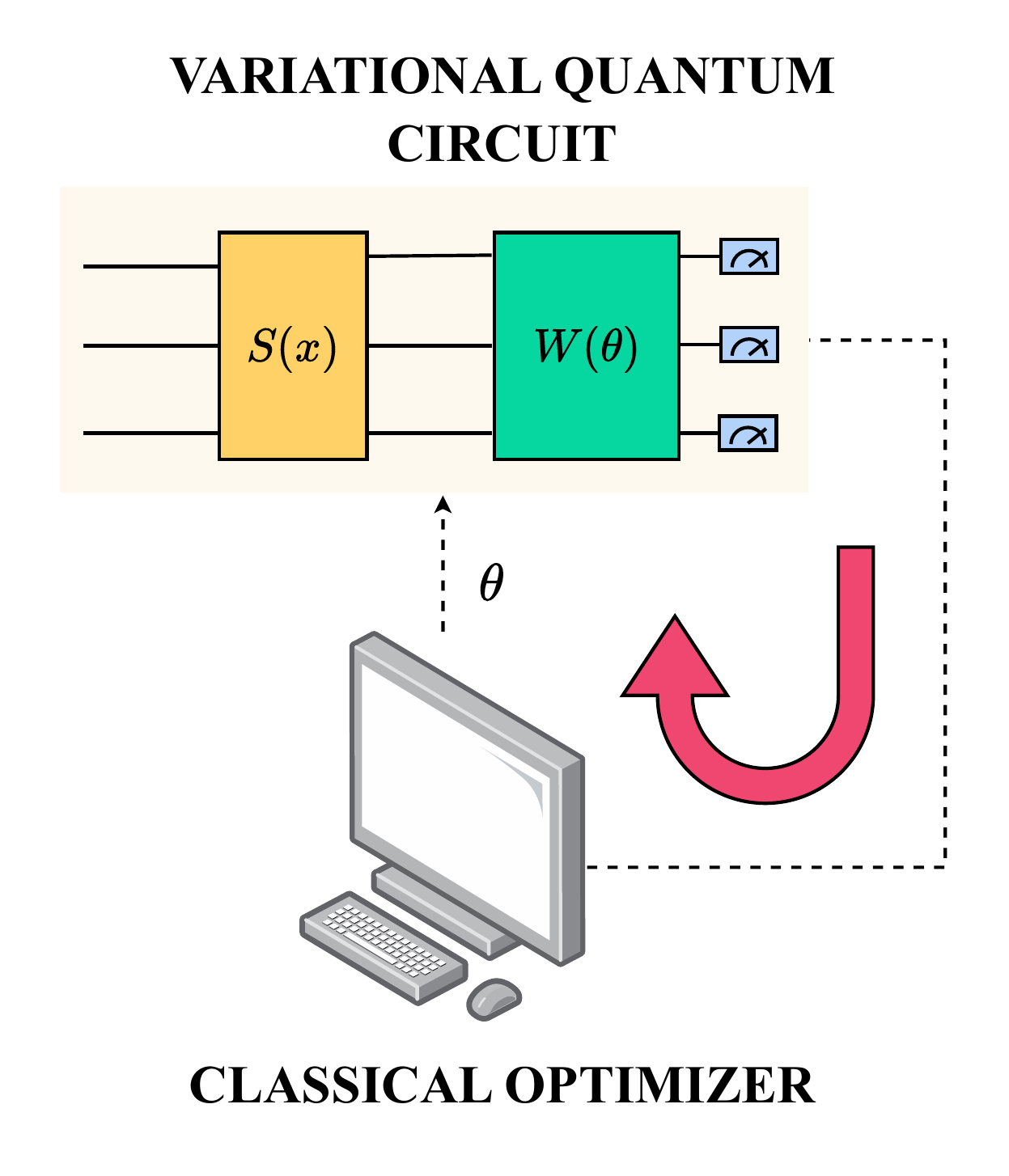} \vspace{-20pt}
    \caption{Quantum Neural Network (QNN). }
    \label{fig:VQC}
\end{figure}
\section{Implementation} \label{sec:Implementation}

This work focuses on analyzing two of the most widely adopted near-term Quantum Machine Learning (QML) models: the \textbf{QNN} and the \textbf{QSVM}. These models are the two most explored in the current NISQ devices era, due to the small circuit depth and low amount of qubits requirements.  The performances of these models are compared by investigating the impact of varying their hyperparameters.

\begin{figure}
    \centering \vspace{-30pt}
    \includegraphics[width=0.35\linewidth]{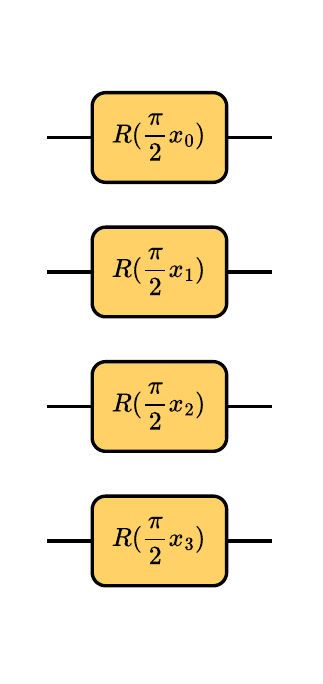} \vspace{-35pt}
    \caption{Representation of a general angle encoding circuit, where R can be any rotational gate. A combination of multiple rotational gates can also be employed. \vspace{-10pt}}
    \label{fig:AngleEncoding}
\end{figure}

For QNN models, which exploit a \textbf{parameterized quantum circuit} (VQC) with parameters optimized during the training phase via a classical optimizer (Figure \ref{fig:VQC}), the impact of the rotational gates used to embed classical data into quantum states through the \textbf{angle encoding} mechanism --- where an example is shown in Figure \ref{fig:AngleEncoding} --- is evaluated. As proved in \cite{tudiscoevaluating2024}, the encoding mechanism serves as a hyperparameter, significantly influencing the overall performance and effectiveness of the model. 

\begin{figure}[h]
    \centering
    \includegraphics[width=0.85\linewidth]{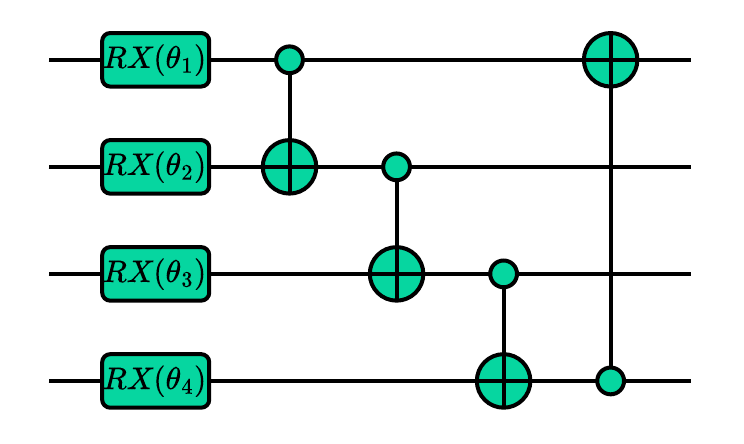}\vspace{-10pt}
    \caption{Example of Basic Entangling Layer circuit. \vspace{-35pt}}
    \label{fig:Basic-Entangling-Layer}
\end{figure}
\begin{figure}[h]
    \centering \vspace{-20pt}
    \includegraphics[width=0.85\linewidth]{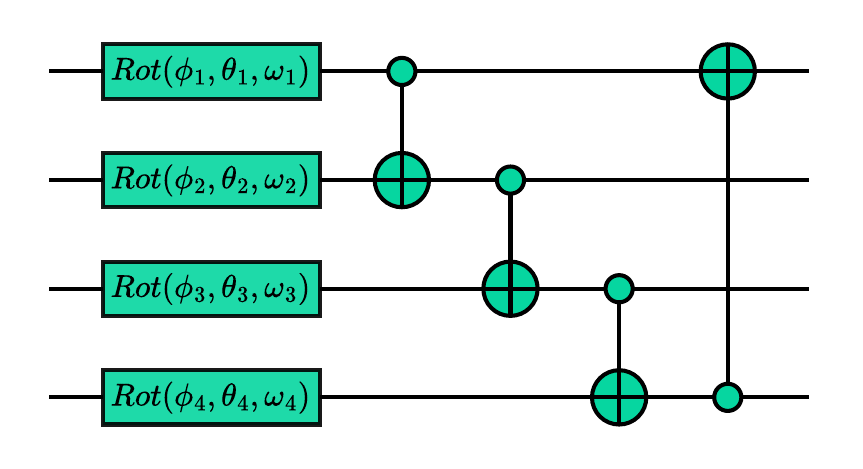} \vspace{-10pt}
    \caption{Example of a Strongly Entangling Layer circuit. \vspace{-1000pt}}
    \label{fig:Strongly-Entangling-Layer}
\end{figure}

Additionally, the \textbf{Strongly Entangling Layer} (Figure \ref{fig:Basic-Entangling-Layer}) and the \textbf{Basic Entangling Layer} (Figure \ref{fig:Strongly-Entangling-Layer}) \textbf{ansatz} are considered, evaluating their expressibility and efficacy in modelling healthcare patterns.

In QNN, the number of measured qubits corresponds to the number of classes in the dataset, where each qubit represents a specific class. Since the raw outputs of the quantum circuit do not inherently represent a probability distribution, a normalization step is required. This is achieved using the softmax function, which transforms the outputs into probabilities --- ensuring that each value lies between 0 and 1, and the sum of all probabilities equals 1. This normalization step allows the model's predictions to be interpreted as class probabilities, enabling accurate and effective classification.

The optimal number of layers of the VQC implementing the QNN is determined by using an iterative approach. 
The process begins with an initial layer count of two, and additional layers are incrementally added --- after each training procedure --- until one of two stopping criteria is met: \begin{enumerate*}
    \item the loss fails to improve for N consecutive iterations in validation, where N is set equal to the number of qubits in the circuit in this work, or
    \item the number of layers exceeds a threshold (set to 100 in this case)
\end{enumerate*}. This approach is analogous to the \textbf{early stopping} technique commonly employed for stopping the training in ML to prevent overfitting and ensure that the model's complexity does not grow unnecessarily.

\begin{figure}[h]
    \centering
    \includegraphics[width=\linewidth]{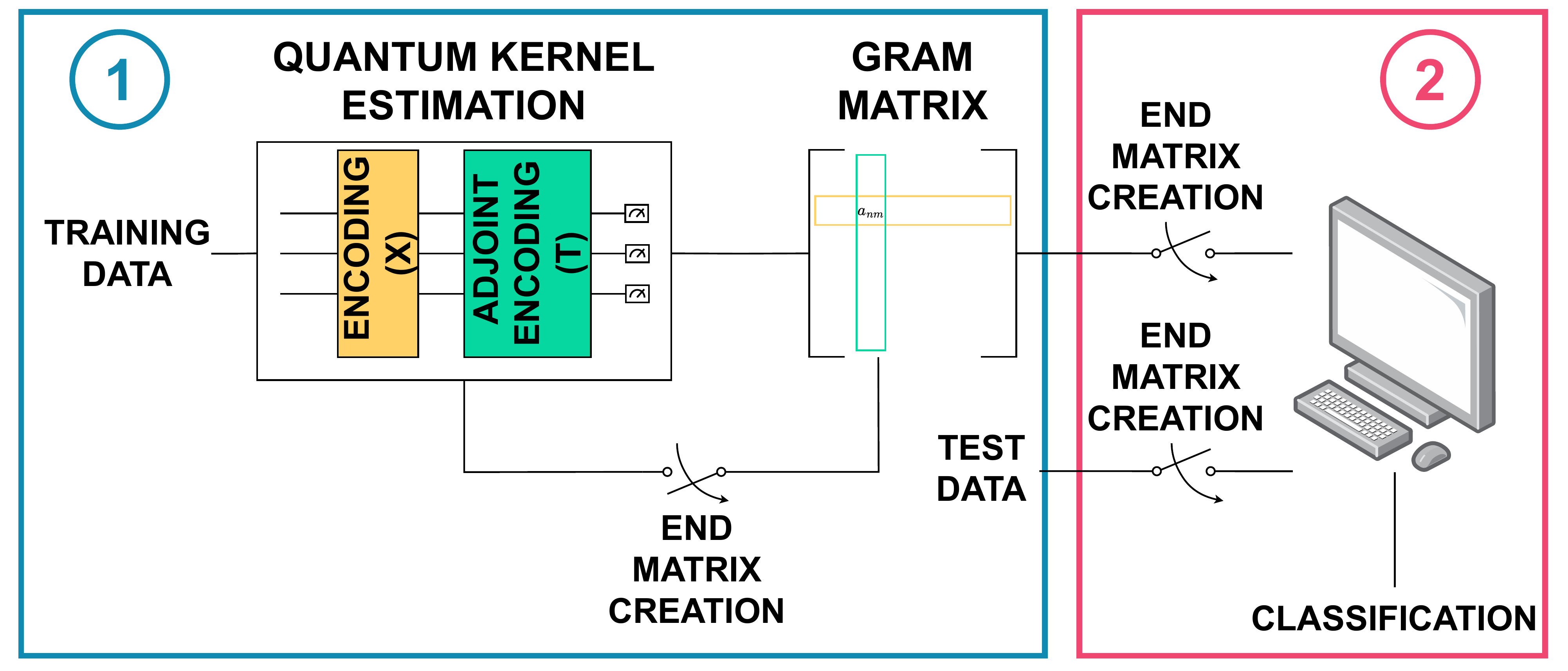} \vspace{-20pt}
    \caption{Quantum Support Vector Machine (QSVM). \vspace{-2000pt}}
    \label{fig:Quantum Kernel estimation}
\end{figure}

\begin{figure}[b]
    \centering \vspace{-20pt}
    \includegraphics[width=0.5\linewidth]{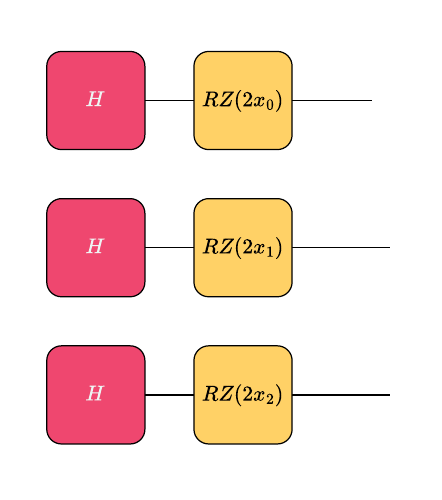} \vspace{-15pt}
    \caption{Z-Feature map. \vspace{-600pt}} 
    \label{fig:Z-feature map}
\end{figure}

On the other hand, the QSVM leverages an encoding circuit followed by a \textbf{successive adjoint encoding circuit} to implement the \textbf{quantum kernel function}, enabling the measurement of similarity between two data points in a high-dimensional Hilbert space, as shown in Figure \ref{fig:Quantum Kernel estimation}. Specifically, this similarity is evaluated using a measurement operator, i.e. a matrix with all zero elements except at position [0,0] --- measuring the probability of reaching the ground state ---, which equals 1 if the encoded values from the two circuits (the encoding and its adjoint) are identical. QSVM effectively identifies relationships between data points by exploiting quantum interference and entanglement, enabling accurate and computationally efficient classification.

\begin{figure}[h]
    \centering
    \includegraphics[width=\linewidth]{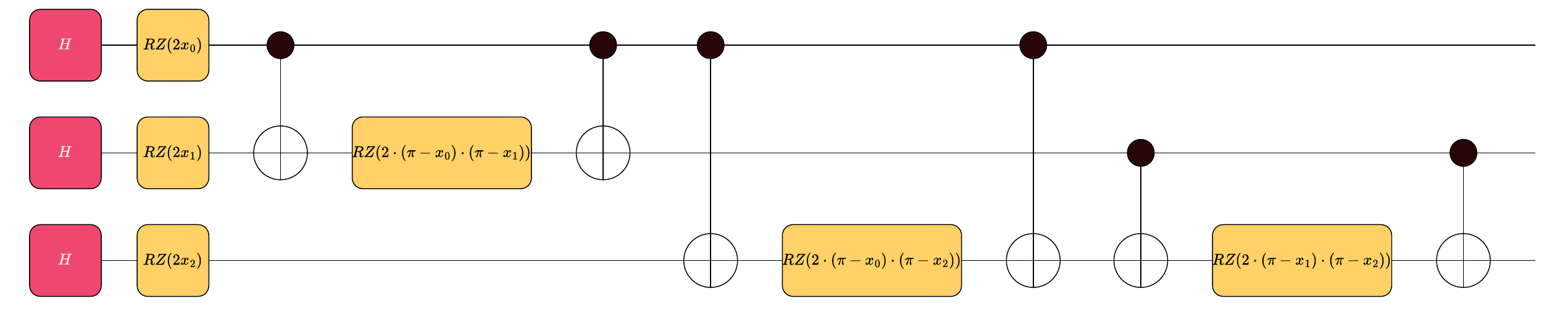} \vspace{-15pt}
    \caption{ZZ-Feature map proposed in \cite{zhou2024quantum}. \vspace{-600pt}}
    \label{fig:ZZ-paper}
\end{figure} \vspace{-20pt}

\begin{figure}[h]
    \centering
    \includegraphics[width=\linewidth]{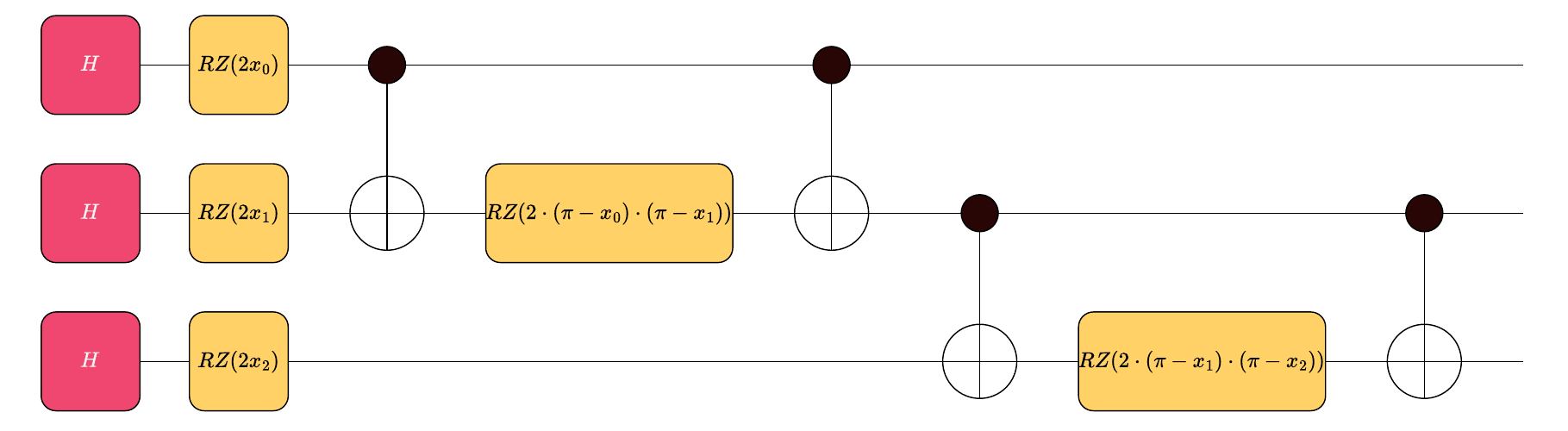} \vspace{-15pt}
    \caption{ZZ-Feature map implemented in Qiskit \cite{zzFeatureMapQiskit}. \vspace{-600pt}}
    \label{fig:ZZ-qiskit}
\end{figure}

The two hyper-parameters investigated for QSVM in this article are the type of encoding, considering the \textbf{Angle}, the \textbf{Z Feature Map} \cite{hossain2021analyzing} (Figure \ref{fig:Z-feature map}), and the two possible variants of \textbf{ZZ Feature Map} --- one presented in \cite{zhou2024quantum} and shown in Figure \ref{fig:ZZ-paper} and the other implemented in Qiskit \cite{zzFeatureMapQiskit} and illustrated in Figure \ref{fig:ZZ-qiskit}  --- and the number of repetitions of encoding and adjoint circuit.

All models considered in this study are trained by applying \textbf{balancing techniques} to address the inherent class imbalance in healthcare datasets. In particular, a \textbf{weighting strategy} is exploited, assigning a higher weight to the minority class with respect to the majority class. This is achieved by scaling the weights: the weight for the minority class is multiplied by the proportion of elements in the majority class, and vice versa. This approach ensures that the training process compensates for the class imbalance, enabling the model to effectively learn from the minority class without being biased by the majority class.  Moreover, it avoids the need for threshold optimization during classification, enhancing the robustness of the prediction mechanism.

Finally, we compare the quantum models with their classical counterparts, considering \textbf{Logistic Regression}, \textbf{Decision Tree}, \textbf{Random Forest}, and \textbf{Support Vector Machine} (\textbf{SVM}). In the case of the SVM, four kernel functions were considered: linear, polynomial (degree 3), radial basis function (RBF), and sigmoid. 
As well as the quantum models, all classical models are trained using a weighting strategy to address the class imbalance in the dataset, ensuring a fair comparison.

 \section{Results} \label{sec:Results} 

 \subsection{Figures of Merit}

Considering that healthcare datasets are often highly imbalanced, in this analysis we have considered figures of merit for comparing the models, focusing on their capability of \textbf{correctly identifying the positive outcomes}, i.e., cancer, \textbf{without exceeding in costly false positive alarms}.  

The first metric is \textbf{Recall} (or sensitivity) and quantifies the ability to detect true positives. It is computed as follows:
\begin{equation}
    \textrm{R} = \frac{\textrm{TP}}{\textrm{TP} +\textrm{FN}} \, ,
    \label{eq:recall}
\end{equation}
where TP is the count of true positives and FN is the count of false negatives.

The second metric is \textbf{Precision}, defined as the proportion of true positives among all predicted positives:
\begin{equation}
    \textrm{P} = \frac{\textrm{TP}}{\textrm{TP} +\textrm{FP}} \, ,
    \label{eq:precision}
\end{equation}
where FP is the count of false positives. 

To balance precision and recall, we use a composite metric called the 	\textbf{F1 score}, which is defined as:
\begin{equation}
    \textrm{F1Score} = \frac{2 \cdot \textrm{R} \cdot \textrm{P}}{\textrm{R} + \textrm{P}} \, .
    \label{eq:F1Score}
\end{equation}
This metric effectively represents a compromise between the need for accurately identifying positive instances and limiting false positives.

\begin{figure}[h]
    \centering
    \includegraphics[width=0.9\linewidth]{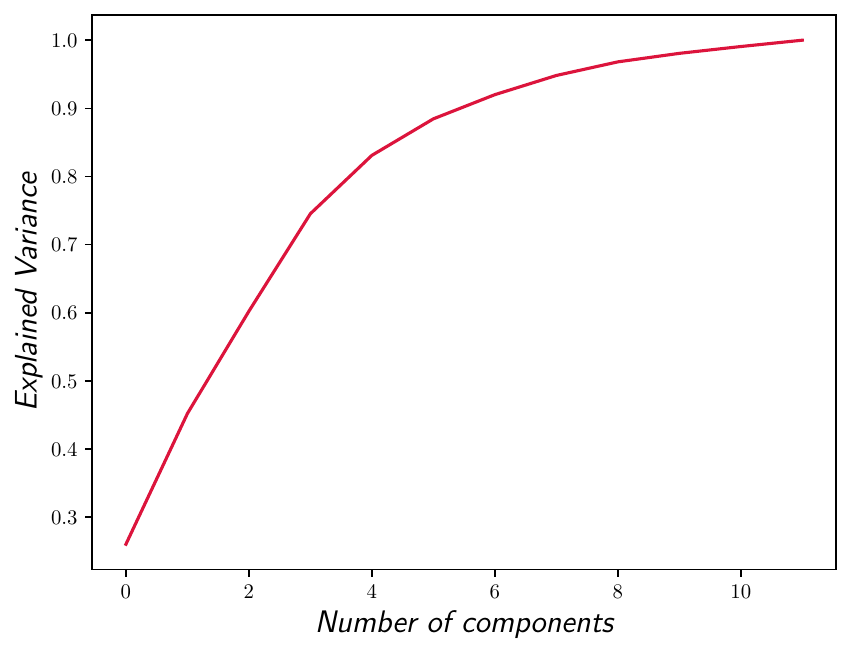} \vspace{-10pt}
    \caption{The cumulative explained variance for each component in Heart Failure dataset, applying Principal Component Analysis. Notice that 5 features include more than 90\% of the variance. \vspace{-5pt}}
    \label{fig:PCA_VarianceHeart}
\end{figure}

\subsection{Datasets Analysis}
In this analysis, we considered the Heart Failure dataset \cite{heart_failure_clinical_records_519}, the Diabetes dataset \cite{diabetes_dataset}, and the Prostate Cancer dataset \cite{prostate_cancer}.

The \textbf{Heart Failure} dataset includes information from 299 patients collected over a follow-up period. Each patient is characterized by 13 features, while the label indicates whether the patient survived (class 0) or died (class 1) during the follow-up period. The 32\% of elements belong to class 1, while the others to class 0.

\begin{figure}[h]
    \centering
    \includegraphics[width=0.9\linewidth]{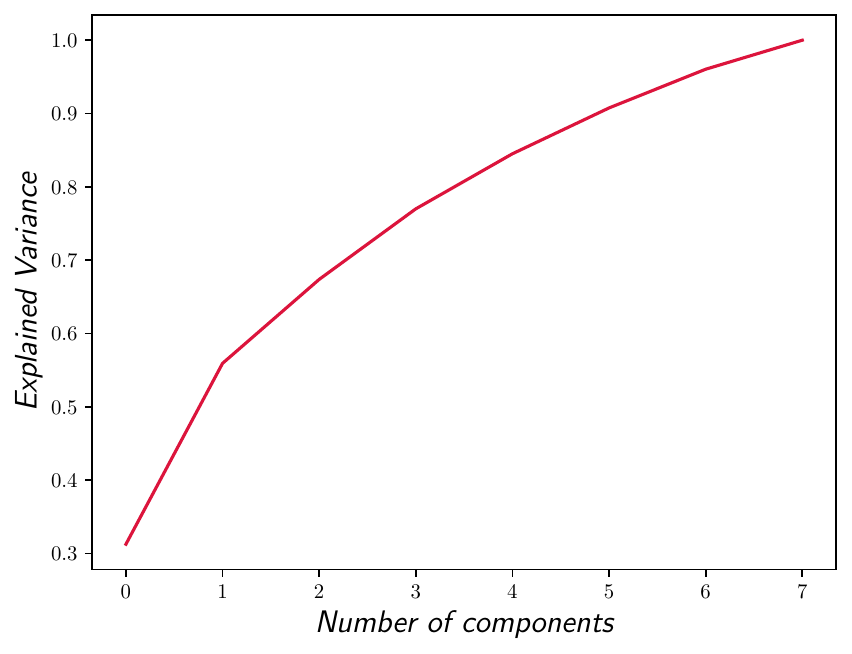}  \vspace{-10pt}
    \caption{The cumulative explained variance for each component in Diabetes dataset, applying Principal Component Analysis. Notice that 6 features include more than 90\% of the variance. \vspace{-5pt} }
    \label{fig:PCA_VarianceDiabetes}
\end{figure}

The \textbf{Diabetes} dataset, originally sourced from the National Institute of Diabetes and Digestive and Kidney Diseases, comprises 768 elements, each characterized by 8 features. The output labels indicate  whether a patient has diabetes based on diagnostic measurements. The 34.9\% of elements belong to class 1 (diabetic), while the others to class (non-diabetic).

\begin{figure}[h]
    \centering
    \includegraphics[width=0.9\linewidth]{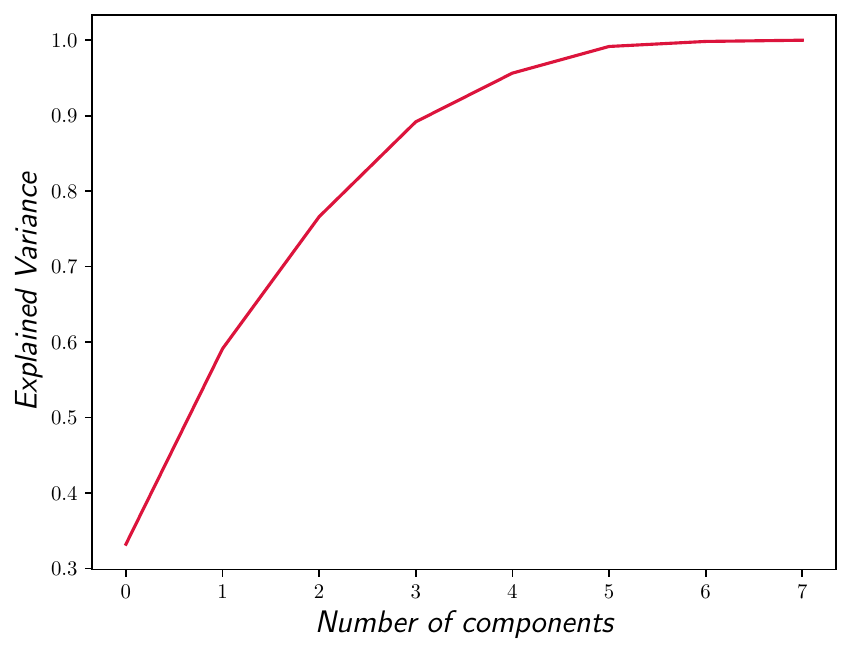}  \vspace{-10pt}
    \caption{The cumulative explained variance for each component in Prostate Cancer dataset, applying Principal Component Analysis. Notice that 6 features include more than 99\% of the variance. \vspace{-10pt} }
    \label{fig:PCA_Varianceprostate}
\end{figure}

Lastly, the \textbf{Prostate Cancer} dataset consists of data from 100 patients, each described by 9 variables. The 2 possible classes are M (malignant) and B (benign). Malignant cases correspond to the 62\% of the elements in the entire data set, while the others are benign. This dataset exhibits the least imbalance among the three.

 \subsection{Performed Tests}

A pre-processing pipeline is applied to all datasets to ensure compatibility with the quantum models. First, each dataset is split into three subsets: training, validation, and test sets. Next, the features are scaled and processed using \textbf{Principal Component Analysis} (\textbf{PCA}) to reduce dimensionality --- in terms of the number of features --- by neglecting components with less information, i.e., those with lower variance.
Finally, the datasets are normalized to the interval (-1,1) to enable their embedding into a quantum state.

Observing Figure \ref{fig:PCA_VarianceHeart} --- which shows the cumulative explained variance ---, it is possible to notice that 5 components capture more than 90\% of the variance. Analogously, in the Prostate Cancer dataset 6 components effectively account for the entire variance, as proved by Figure \ref{fig:PCA_Varianceprostate}. Conversely, in the Diabetes dataset, each component contributes significantly to the overall variance (Figure \ref{fig:PCA_VarianceDiabetes}), making feature reduction without performance loss more challenging. However, 6 components are sufficient to express approximately 90\% of the variance. 

The models were trained for 100 epochs with early stopping, using a patience parameter of 5. To address the previously mentioned class imbalance, the \textbf{cross-entropy loss function} was used with \textbf{class weights adjusted} according to the sample distribution in the training set as discussed in Section \ref{sec:Implementation}.

After training all the models, those with an F1-score below a predefined threshold on the training set are discarded. Then, the model configurations achieving the highest F1-score on the validation set were selected. The initial filtering based on the training set limits the possibility of selecting a model that performs well on the validation set due to overfitting but exhibits poor performance on the test set and, consequently, on new samples.

{\renewcommand{\arraystretch}{1.2}
\begin{table}[h]\vspace{-10pt}
 \centering
 \caption{ Precision (P) and Recall (R) obtained by QNN on the Heart-Failure dataset.}
 \begin{widetable}{\columnwidth}{ccccccccccc}
 \hline
 \multirow{2}{*}{\textbf{Feat}} & \multirow{2}{*}{\textbf{Enc}} & \multirow{2}{*}{\textbf{Reup}} & \multirow{2}{*}{\textbf{Ansatz}} & \multirow{2}{*}{\textbf{Layers}} & \multicolumn{2}{c}{\textbf{Train}} & \multicolumn{2}{c}{\textbf{Validation}} & \multicolumn{2}{c}{\textbf{Test}}\\
 & & & & & \textbf{P} & \textbf{R} & \textbf{P} & \textbf{R} & \textbf{P} & \textbf{R} \\
 \hline
2 & XZY & True & strongly &2 &0.40 & 0.68 & 0.44 & 0.73 & 0.33 & 0.58 \\
3 & YX & True & strongly &6 &0.53 & 0.66 & 0.61 & 0.73 & 0.28 & 0.26 \\
4 & X & True & strongly &2 &0.43 & 0.77 & 0.52 & 0.80 & 0.21 & 0.37 \\
5 & YXZ & True & basic &13 &0.48 & 0.73 & 0.60 & 0.80 & 0.27 & 0.37 \\
 \hline
 \end{widetable}
 \label{tab:Heart-failureVQC}
 \end{table}
}
{\renewcommand{\arraystretch}{1.2}
\begin{table}[h] \vspace{-10pt}
 \centering
 \caption{Precision (P) and Recall (R) obtained by QSVM on the Heart-Failure dataset.}
 \begin{widetable}{\columnwidth}{ccccccccc}
 \hline
 \multirow{2}{*}{\textbf{Feat}} & \multirow{2}{*}{\textbf{Enc}} & \multirow{2}{*}{\textbf{Rep}} & \multicolumn{2}{c}{\textbf{Train}} & \multicolumn{2}{c}{\textbf{Validation}} & \multicolumn{2}{c}{\textbf{Test}}\\
 & & & \textbf{P} & \textbf{R} & \textbf{P} & \textbf{R} & \textbf{P} & \textbf{R} \\
 \hline
2 & ZZ-qiskit & 2 &0.44 & 0.74 & 0.43 & 0.67 & 0.30 & 0.53 \\
3 & Angle & 3 &0.46 & 0.68 & 0.57 & 0.80 & 0.21 & 0.32 \\
4 & Angle & 3 &0.46 & 0.74 & 0.48 & 0.80 & 0.27 & 0.37 \\
5 & ZZ & 3 &0.81 & 0.94 & 0.71 & 0.67 & 0.33 & 0.21 \\
 \hline
 \end{widetable}
 \label{tab:Heart-failureQSVM} 
 \end{table}
}
\subsubsection{Heart-Failure}
Tables \ref{tab:Heart-failureVQC} and \ref{tab:Heart-failureQSVM} present the results achieved in terms of Precision and Recall by the best-performing QNN and QSVM models on the Heart-Failure dataset, by varying numbers of features considered. It is possible to notice that the QNN models generally achieve higher Recall on the training and validation sets, which is advantageous for applications where minimizing false negatives is critical. On the other hand, QSVM models show slightly more balanced performance across Precision and Recall, particularly in configurations with fewer features. 

\begin{figure}[h]
    \centering
    \includegraphics[width=\linewidth]{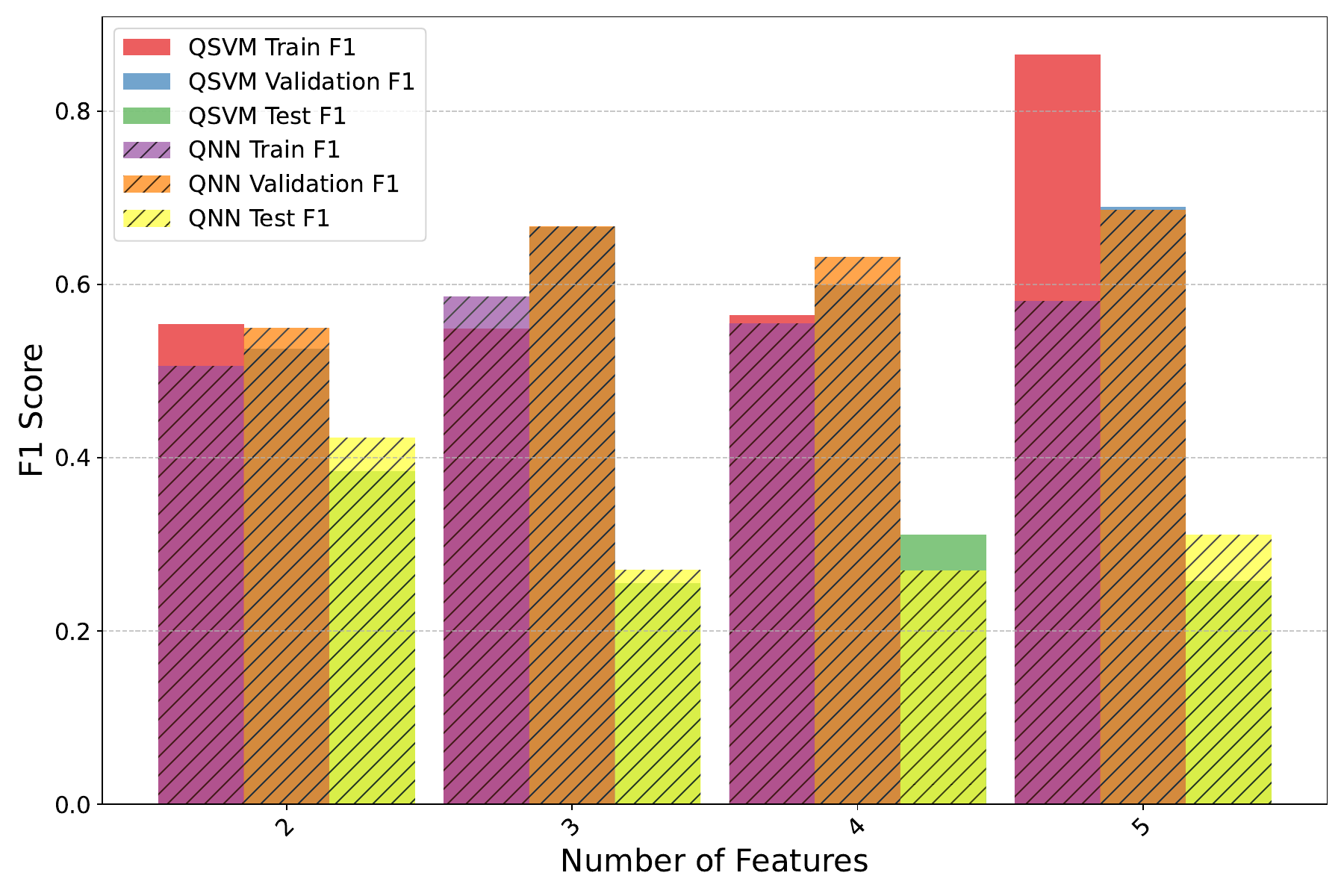} \vspace{-10pt}
    \caption{Results of the QNN and QSVM models on the Heart-Failure dataset changing the number of features. For the QNN, the models accepted must have a F1-score on the training set higher than the 50\%.}
    \label{fig:heart-failure}  
\end{figure}

{\renewcommand{\arraystretch}{1.2}
\begin{table}[h]  \vspace{-10pt}
 \centering
 \caption{Precision (P) and Recall (R) obtained by classical models on the Heart-Failure dataset.}
 \begin{widetable}{\columnwidth}{ccccccccc}
 \hline
 \multirow{2}{*}{\textbf{Feat}} & \multirow{2}{*}{\textbf{Model}} & \multirow{2}{*}{\textbf{Kernel}} & \multicolumn{2}{c}{\textbf{Train}} & \multicolumn{2}{c}{\textbf{Validation}} & \multicolumn{2}{c}{\textbf{Test}}\\
 & & & \textbf{P} & \textbf{R} & \textbf{P} & \textbf{R} & \textbf{P} & \textbf{R} \\
 \hline
2 & LogisticRegression & - &0.35 & 0.55 & 0.47 & 0.60 & 0.30 & 0.58 \\
3 & svm & rbf &0.43 & 0.66 & 0.47 & 0.60 & 0.26 & 0.47 \\
4 & DecisionTree & - &1.00 & 1.00 & 0.54 & 0.47 & 0.33 & 0.16 \\
5 & svm & poly &0.46 & 0.71 & 0.50 & 0.80 & 0.20 & 0.26 \\
 \hline
 \end{widetable} \vspace{-10pt}
 \label{tab:Heart-failureClassic}
 \end{table}
}

{\renewcommand{\arraystretch}{1.2}
\begin{table}[b] \vspace{-10pt}
 \centering
 \caption{Comparison of classical and quantum models in terms of Precision (P) and Recall (R) on the test set for Heart-Failure dataset.}
 \begin{widetable}{\columnwidth}{ccccccc}
 \hline
 \multirow{2}{*}{\textbf{Feat}} & \multicolumn{2}{c}{\textbf{QNN}} & \multicolumn{2}{c}{\textbf{QSVM}} & \multicolumn{2}{c}{\textbf{Classical}}\\
 & \textbf{P} & \textbf{R} & \textbf{P} & \textbf{R} & \textbf{P} & \textbf{R} \\
 \hline
2 &0.33 & \textbf{0.58} & \textbf{0.30} & 0.53 & \textbf{0.30} & \textbf{0.58} \\
3 &\textbf{0.28} & 0.26 & 0.21 & 0.32 & 0.26 & \textbf{0.47} \\
4 &0.21 & \textbf{0.37} & 0.27 & \textbf{0.37} & \textbf{0.33} & 0.16 \\
5 &0.27 & \textbf{0.37} & \textbf{0.33} & 0.21 & 0.20 & 0.26 \\
 \hline
 \end{widetable}
 \label{tab:comparisonHeart-failure}
 \end{table}
}

The two quantum models are also compared in terms of F1-score for Heart-failure dataset in Figure \ref{fig:heart-failure}. It shows that QNN models outperform QSVM models in test performance, achieving higher F1-scores except when using 4 features. This indicates that QNN models are more effective in generalizing unseen data in this specific application.
\begin{figure}[h]
    \centering
    \includegraphics[width=\linewidth]{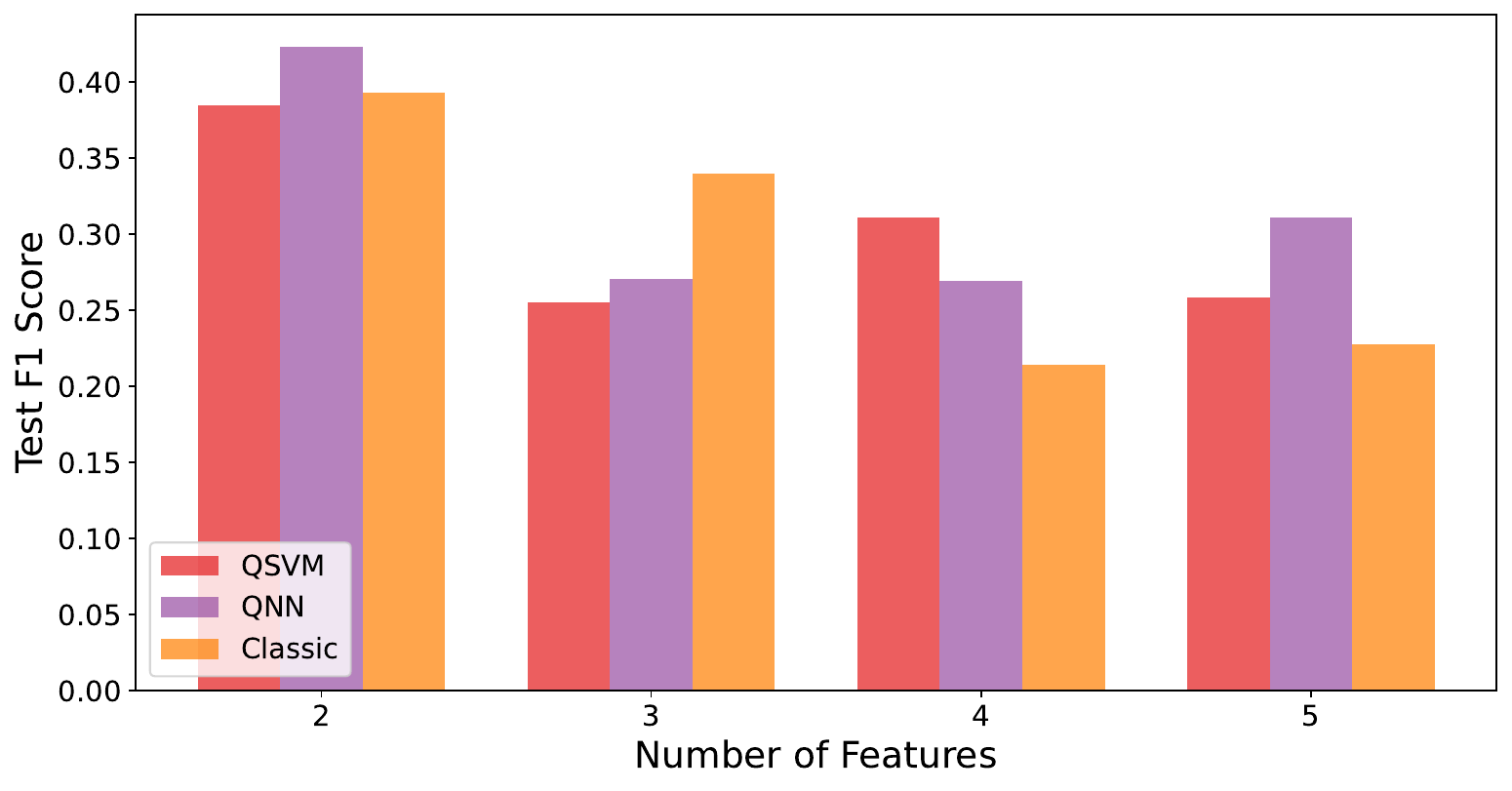} \vspace{-20pt}
    \caption{Comparison of the best Quantum Models with respect to the best classical on the F1-score on the test set of the Heart Failure dataset.}
    \label{fig:HeartFailure_classical}
\end{figure}

Moreover, the quantum models are also compared with the best classical ones, whose results and configuration are reported in Table \ref{tab:Heart-failureClassic}. As the number of features increases, the classical models exhibit overfitting, achieving strong performance on the training and validation sets but performing poorly on the test set.  As shown in Table \ref{tab:comparisonHeart-failure} and in Figure \ref{fig:HeartFailure_classical}, the classical models achieve precision on the test set comparable to that of QNN and QSVM. However, their recall declines significantly as the number of features increases, highlighting challenges in maintaining sensitivity with larger feature sets and the advantage of quantum models, which demonstrate greater robustness in recall across different feature configurations.

\subsubsection{Diabetes}
{\renewcommand{\arraystretch}{1.2}
\begin{table}[h] \vspace{-10pt}
 \centering
 \caption{Precision (P) and Recall (R) obtained by QNN on the Diabetes dataset.}
 \begin{widetable}{\columnwidth}{ccccccccccc}
 \hline
 \multirow{2}{*}{\textbf{Feat}} & \multirow{2}{*}{\textbf{Enc}} & \multirow{2}{*}{\textbf{Reup}} & \multirow{2}{*}{\textbf{Ansatz}} & \multirow{2}{*}{\textbf{Layers}} & \multicolumn{2}{c}{\textbf{Train}} & \multicolumn{2}{c}{\textbf{Validation}} & \multicolumn{2}{c}{\textbf{Test}}\\
 & & & & & \textbf{P} & \textbf{R} & \textbf{P} & \textbf{R} & \textbf{P} & \textbf{R} \\
 \hline
2 & XYZ & True & strongly &4 &0.59 & 0.75 & 0.58 & 0.60 & 0.51 & 0.57 \\
3 & YZX & True & strongly &7 &0.58 & 0.77 & 0.56 & 0.79 & 0.58 & 0.81 \\
4 & XYZ & False & strongly &3 &0.62 & 0.74 & 0.61 & 0.79 & 0.54 & 0.69 \\
5 & XYZ & True & strongly &6 &0.57 & 0.76 & 0.67 & 0.79 & 0.53 & 0.59 \\
6 & Y & True & strongly &25 &0.65 & 0.84 & 0.57 & 0.74 & 0.59 & 0.69 \\
 \hline
 \end{widetable}
 \label{tab:diabetesVQC}
 \end{table}
}
{\renewcommand{\arraystretch}{1.2}
\begin{table}[h]
 \centering
 \caption{Precision (P) and Recall (R) obtained by QSVM on the Diabetes dataset.}
 \begin{widetable}{\columnwidth}{ccccccccc}
 \hline
 \multirow{2}{*}{\textbf{Feat}} & \multirow{2}{*}{\textbf{Enc}} & \multirow{2}{*}{\textbf{Rep}} & \multicolumn{2}{c}{\textbf{Train}} & \multicolumn{2}{c}{\textbf{Validation}} & \multicolumn{2}{c}{\textbf{Test}}\\
 & & & \textbf{P} & \textbf{R} & \textbf{P} & \textbf{R} & \textbf{P} & \textbf{R} \\
 \hline
2 & Z & 3 &0.56 & 0.75 & 0.56 & 0.67 & 0.48 & 0.57 \\
3 & ZZ & 1 &0.56 & 0.76 & 0.55 & 0.84 & 0.59 & 0.76 \\
4 & Angle & 3 &0.61 & 0.79 & 0.58 & 0.81 & 0.56 & 0.74 \\
5 & Angle & 2 &0.56 & 0.75 & 0.68 & 0.79 & 0.58 & 0.78 \\
6 & Z & 2 &0.62 & 0.86 & 0.55 & 0.77 & 0.64 & 0.83 \\
 \hline
 \end{widetable}
 \label{tab:diabetesQSVM}
 \end{table}
}

Tables \ref{tab:diabetesVQC} and \ref{tab:diabetesQSVM} present the results achieved in terms of Precision and Recall by the best-performing QNN and QSVM models on the Diabetes dataset, by varying numbers of features considered. QSVM outperforms QNN in recall with higher feature counts. On the other hand, QNN provides a more balanced Precision and Recall, which can be advantageous in the 2 feature cases.  In general, both models achieve their best test performance with 6 features, indicating this as an optimal feature set for the Diabetes dataset under the given configurations. This is reasonable with expectation since, as observed from Figure \ref{fig:PCA_VarianceDiabetes},  each component introduces significant variability, making configurations with fewer features less effective.

\begin{figure}[h]
    \centering
    \includegraphics[width=\linewidth]{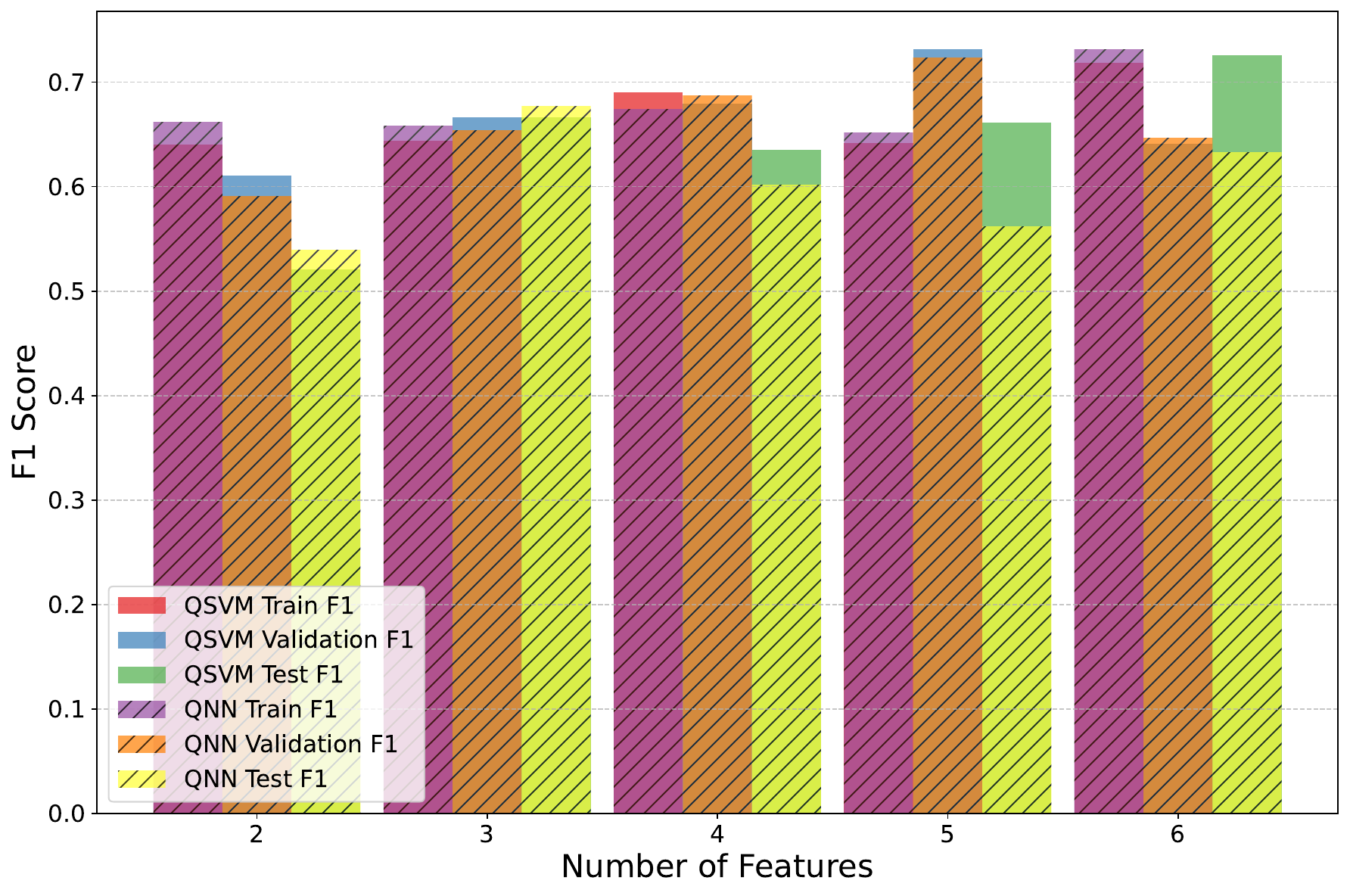} \vspace{-10pt}
    \caption{Results of the QNN and QSVM models on the Diabetes dataset compared changing the number of features. For the QNN, the models accepted must have a F1-score on the training set higher than the 65\%. }
    \label{fig:diabetes-cancer}
\end{figure}

The two quantum models are also evaluated in terms of F1-score for the Diabetes dataset in Figure \ref{fig:diabetes-cancer}. QSVM models outperform QSVM models in test performance, achieving higher F1-scores except when using 2 and 3 features. 

{\renewcommand{\arraystretch}{1.2}
\begin{table}[h]
 \centering
 \caption{Precision (P) and Recall (R) obtained by classical models on the Diabetes dataset.}
 \begin{widetable}{\columnwidth}{ccccccccc}
 \hline
 \multirow{2}{*}{\textbf{Feat}} & \multirow{2}{*}{\textbf{Model}} & \multirow{2}{*}{\textbf{Kernel}} & \multicolumn{2}{c}{\textbf{Train}} & \multicolumn{2}{c}{\textbf{Validation}} & \multicolumn{2}{c}{\textbf{Test}}\\
 & & & \textbf{P} & \textbf{R} & \textbf{P} & \textbf{R} & \textbf{P} & \textbf{R} \\
 \hline
2 & svm & rbf &0.57 & 0.80 & 0.53 & 0.60 & 0.52 & 0.70 \\
3 & svm & rbf &0.56 & 0.78 & 0.54 & 0.81 & 0.55 & 0.78 \\
4 & svm & rbf &0.59 & 0.81 & 0.58 & 0.81 & 0.58 & 0.76 \\
5 & svm & linear &0.57 & 0.68 & 0.72 & 0.72 & 0.53 & 0.63 \\
6 & svm & rbf &0.62 & 0.87 & 0.55 & 0.77 & 0.64 & 0.81 \\
 \hline
 \end{widetable}
 \label{tab:diabetesClassic}
 \end{table}
}

{\renewcommand{\arraystretch}{1.2}
\begin{table} [h] \vspace{-10pt}
 \centering
 \caption{Comparison of classical and quantum models in terms of Precision (P) and Recall (R) on test set for Diabetes dataset.}
 \begin{widetable}{\columnwidth}{ccccccc}
 \hline
 \multirow{2}{*}{\textbf{Feat}} & \multicolumn{2}{c}{\textbf{QNN}} & \multicolumn{2}{c}{\textbf{QSVM}} & \multicolumn{2}{c}{\textbf{Classical}}\\
 & \textbf{P} & \textbf{R} & \textbf{P} & \textbf{R} & \textbf{P} & \textbf{R} \\
 \hline
2 &0.51 & 0.57 & 0.48 & 0.57 & \textbf{0.52} & \textbf{0.70} \\
3 &0.58 & \textbf{0.81} & \textbf{0.59} & 0.76 & 0.55 & 0.78 \\
4 &0.54 & 0.69 & 0.56 & 0.74 & \textbf{0.58} & \textbf{0.76} \\
5 &0.53 & 0.59 & \textbf{0.58} & \textbf{0.78} & 0.53 & 0.63 \\
6 &0.59 & 0.69 & \textbf{0.64} & \textbf{0.83} & \textbf{0.64 }& 0.81 \\
 \hline
 \end{widetable}
 \label{tab:comparisonDiabetes}
 \end{table} \vspace{-10pt}
}

\begin{figure}[h]
    \centering
    \includegraphics[width=\linewidth]{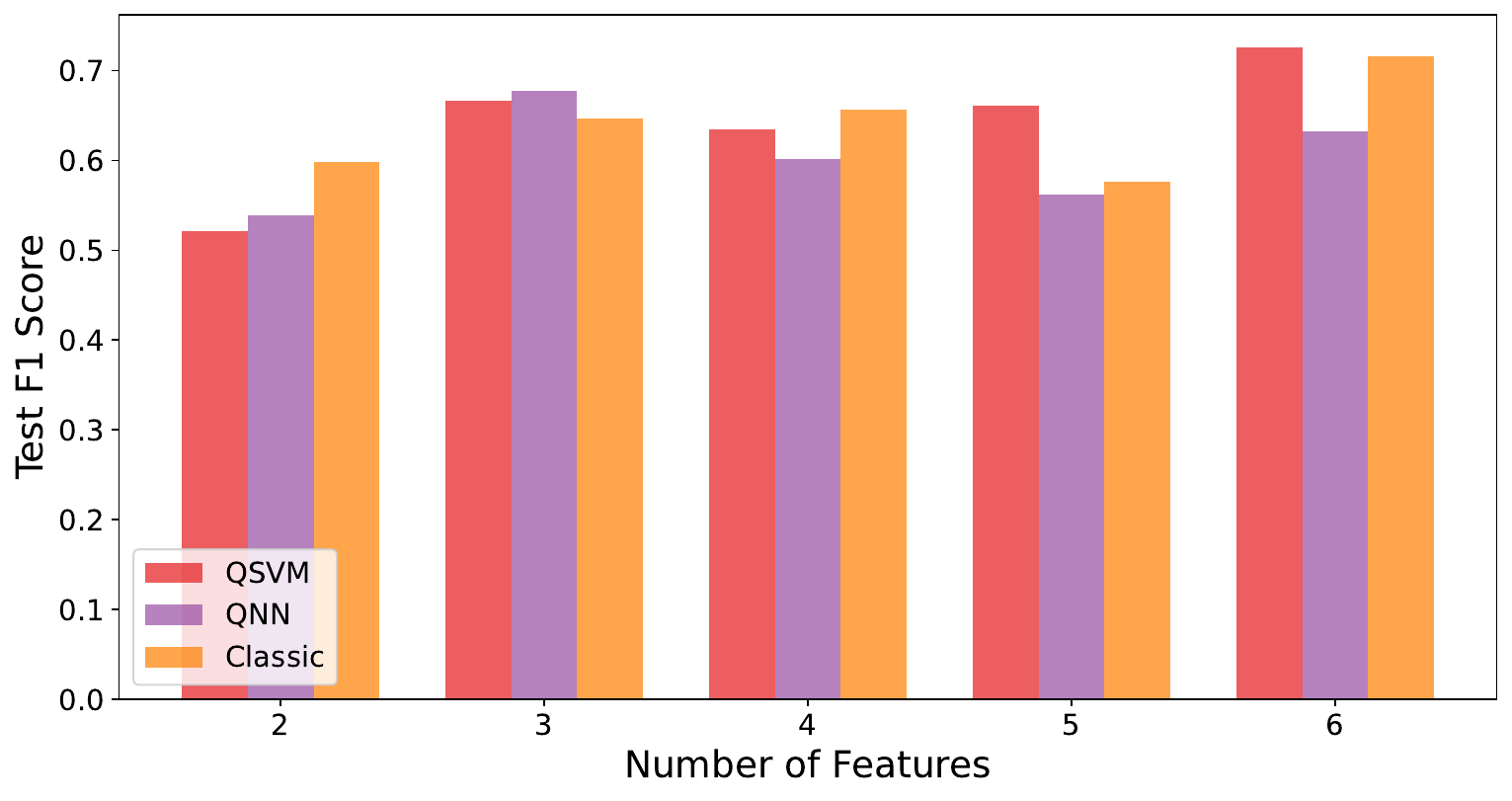}
    \caption{Comparison of the best Quantum Models with respect to the best classical on the F1-score on the test set of the Diabetes dataset.}
    \label{fig:Diabetes_classical}
\end{figure}

The best classical models among those considered were also identified and reported in Table \ref{tab:diabetesClassic}. For this dataset, the best-performing classical model is almost always the SVM with the radial basis function. In some cases, such as with five features, overfitting can be observed as the Precision and Recall in validation are significantly higher than those in the test set. As shown in Table \ref{tab:comparisonDiabetes}, the classical models achieve the best performance, in particular highest precision for 2 and 4 features. QNN gives the best Recall in the 3 features case, while QSVM outperforms the other models with 5 and 6 features, especially in terms of Recall, showing better sensitivity in identifying the positives. These conclusions can also be drawn by comparing quantum and classical models in terms of F1-score, as shown in Figure \ref{fig:Diabetes_classical}.

\subsubsection{Prostate Cancer}
{\renewcommand{\arraystretch}{1.2}
\begin{table}[h] \vspace{-10pt}
 \centering
 \caption{Precision (P) and Recall (R) obtained by QNN on the Prostate Cancer dataset.}
 \begin{widetable}{\columnwidth}{ccccccccccc}
 \hline
 \multirow{2}{*}{\textbf{Feat}} & \multirow{2}{*}{\textbf{Enc}} & \multirow{2}{*}{\textbf{Reup}} & \multirow{2}{*}{\textbf{Ansatz}} & \multirow{2}{*}{\textbf{Layers}} & \multicolumn{2}{c}{\textbf{Train}} & \multicolumn{2}{c}{\textbf{Validation}} & \multicolumn{2}{c}{\textbf{Test}}\\
 & & & & & \textbf{P} & \textbf{R} & \textbf{P} & \textbf{R} & \textbf{P} & \textbf{R} \\
 \hline
2 & XZY & True & strongly &6 &0.74 & 0.78 & 0.75 & 0.90 & 0.67 & 0.50 \\
3 & X & True & basic &2 &0.74 & 0.78 & 0.82 & 0.90 & 0.82 & 0.75 \\
4 & X & True & basic &6 &0.72 & 0.82 & 1.00 & 0.90 & 0.75 & 0.50 \\
5 & YX & False & basic &20 &0.81 & 0.85 & 0.91 & 1.00 & 0.91 & 0.83 \\
6 & YZX & False & basic &8 &0.71 & 0.80 & 0.83 & 1.00 & 0.73 & 0.67 \\
 \hline
 \end{widetable}
 \label{tab:ProstateVQC}
 \end{table}
}
{\renewcommand{\arraystretch}{1.2}
\begin{table}[h]\vspace{-10pt}
 \centering
 \caption{Precision (P) and Recall (R) obtained by QSVM on the Prostate Cancer dataset.}
 \begin{widetable}{\columnwidth}{ccccccccc}
 \hline
 \multirow{2}{*}{\textbf{Feat}} & \multirow{2}{*}{\textbf{Enc}} & \multirow{2}{*}{\textbf{Rep}} & \multicolumn{2}{c}{\textbf{Train}} & \multicolumn{2}{c}{\textbf{Validation}} & \multicolumn{2}{c}{\textbf{Test}}\\
 & & & \textbf{P} & \textbf{R} & \textbf{P} & \textbf{R} & \textbf{P} & \textbf{R} \\
 \hline
2 & Z & 2 &0.71 & 0.80 & 0.80 & 0.80 & 0.79 & 0.92 \\
3 & Z & 3 &0.82 & 0.82 & 0.67 & 0.80 & 0.70 & 0.58 \\
4 & Angle & 3 &0.90 & 0.90 & 0.80 & 0.80 & 0.85 & 0.92 \\
5 & Angle & 3 &0.95 & 0.90 & 0.89 & 0.80 & 0.82 & 0.75 \\
6 & ZZ & 1 &1.00 & 1.00 & 0.71 & 1.00 & 0.59 & 0.83 \\
 \hline
 \end{widetable}
 \label{tab:ProstateQSVM}
 \end{table}
}

Tables \ref{tab:ProstateVQC} and \ref{tab:ProstateQSVM} show the results obtained in terms of Precision and Recall by the best-performing QNN and QSVM models on the Prostate Cancer dataset, by varying numbers of features considered. QSVM outperforms QNN in terms of Recall in the test set in all cases except when the 3 features are considered. Moreover, QSVM generally provides also better precision with respect to QNN for 2 and 4 features. The relatively poor performance in terms of Recall of QNN can be partially attributed to overfitting as evidenced by significantly higher validation metrics compared to those on the test set.

\begin{figure}[h]
    \centering
    \includegraphics[width=\linewidth]{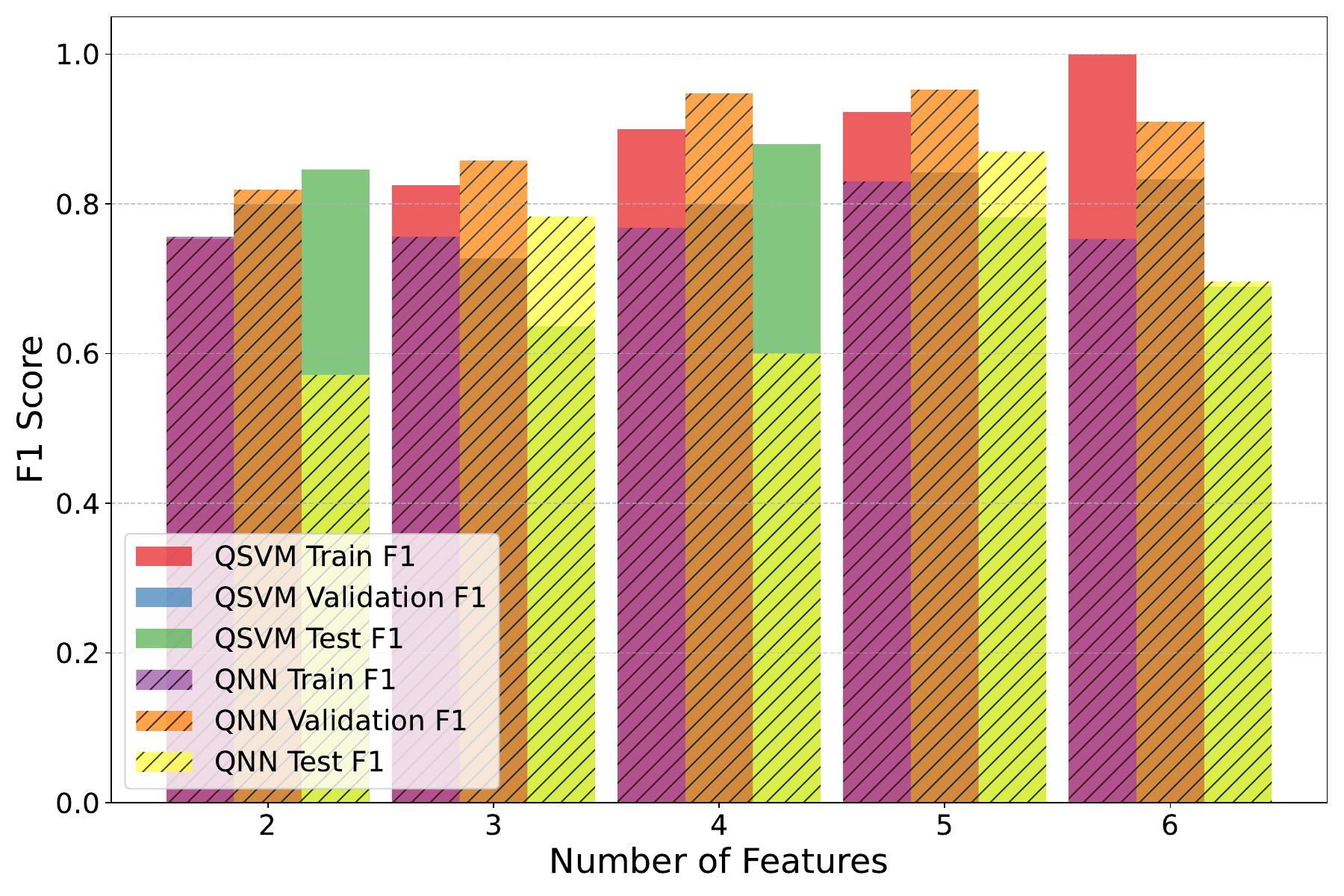} \vspace{-10pt}
    \caption{Results of the QNN and QSVM models on the Prostate Cancer dataset compared changing the number of features. For the QNN, the models accepted must have a F1-score on the training set higher than the 75\%. \vspace{-10pt}}
    \label{fig:prostate-cancer}
\end{figure}

The two quantum models are also compared in terms of F1-score for the Prostate Cancer dataset in Figure \ref{fig:prostate-cancer}. It shows that QNN models outperform QSVM models in test performance, achieving higher F1-scores except when using 2 and 4 features, while QNN is better for 3 and 5 features. For 6 features, the two models exhibit comparable performance.

{\renewcommand{\arraystretch}{1.2}
\begin{table}[h] \vspace{-10pt}
 \centering
 \caption{Precision (P) and Recall (R) obtained by classical models on the Prostate Cancer dataset.}
 \begin{widetable}{\columnwidth}{ccccccccc}
 \hline
 \multirow{2}{*}{\textbf{Feat}} & \multirow{2}{*}{\textbf{Model}} & \multirow{2}{*}{\textbf{Kernel}} & \multicolumn{2}{c}{\textbf{Train}} & \multicolumn{2}{c}{\textbf{Validation}} & \multicolumn{2}{c}{\textbf{Test}}\\
 & & & \textbf{P} & \textbf{R} & \textbf{P} & \textbf{R} & \textbf{P} & \textbf{R} \\
 \hline
2 & RandomForest & - &1.00 & 1.00 & 0.78 & 0.70 & 0.92 & 0.92 \\
3 & RandomForest & - &1.00 & 1.00 & 0.82 & 0.90 & 0.83 & 0.83 \\
4 & RandomForest & - &1.00 & 1.00 & 0.82 & 0.90 & 0.83 & 0.83 \\
5 & DecisionTree & - &1.00 & 1.00 & 0.80 & 0.80 & 0.91 & 0.83 \\
6 & RandomForest & - &1.00 & 1.00 & 0.82 & 0.90 & 0.91 & 0.83 \\
 \hline
 \end{widetable}
 \label{tab:ProstateClassic}
 \end{table}
}
{\renewcommand{\arraystretch}{1.2}
\begin{table}[h]
 \centering
 \caption{Comparison of classical and quantum models in terms of Precision (P) and Recall (R) on the test set for Prostate Cancer dataset.}
 \begin{widetable}{\columnwidth}{ccccccc}
 \hline
 \multirow{2}{*}{\textbf{Feat}} & \multicolumn{2}{c}{\textbf{QNN}} & \multicolumn{2}{c}{\textbf{QSVM}} & \multicolumn{2}{c}{\textbf{Classical}}\\
 & \textbf{P} & \textbf{R} & \textbf{P} & \textbf{R} & \textbf{P} & \textbf{R} \\
 \hline
2 &0.67 & 0.50 & 0.79 & \textbf{0.92} & \textbf{0.92} & \textbf{0.92} \\
3 &0.82 & 0.75 & 0.70 & 0.58 & \textbf{0.83} & \textbf{0.83} \\
4 &0.75 & 0.50 & \textbf{0.85} & \textbf{0.92 }& 0.83 & 0.83 \\
5 &\textbf{0.91} & \textbf{0.83} & 0.82 & 0.75 & \textbf{0.91} & \textbf{0.83} \\
6 &0.73 & 0.67 & 0.59 & \textbf{0.83} & \textbf{0.91} & \textbf{0.83} \\
 \hline
 \end{widetable}
 \label{tab:comparisonProstate}
 \end{table}
}
\begin{figure}[b]
    \centering
    \includegraphics[width=\linewidth]{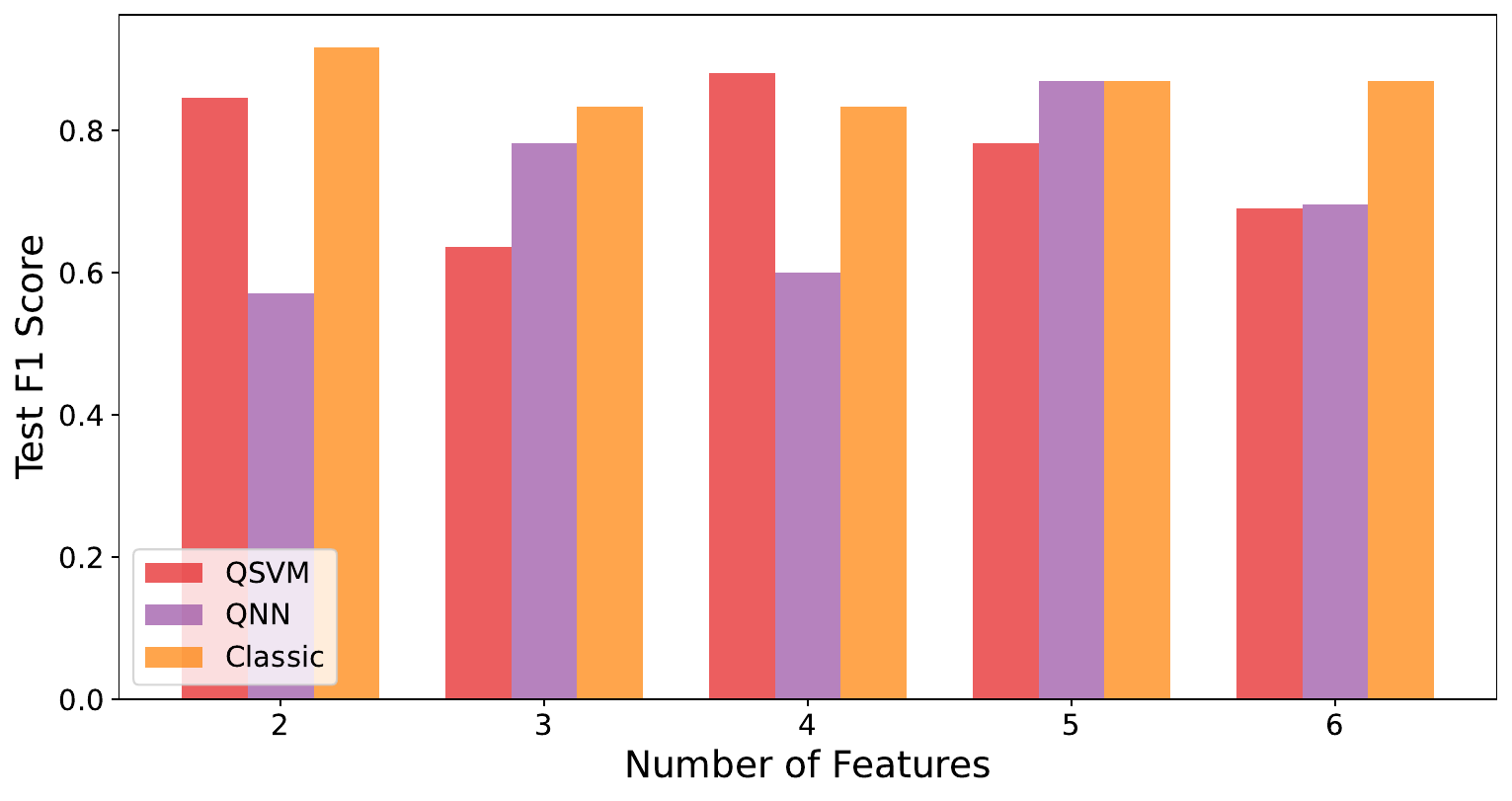}
    \caption{Comparison of the best Quantum Models with respect to the best classical on the F1-score on the test set of the Prostate Cancer dataset.}
    \label{fig:Prostate_classical}
\end{figure}

The best classical models among those considered were also identified and reported in Table \ref{tab:ProstateClassic}. Random Forest results in general as the best performing classical model among those considered in this work. For this dataset, classical models perform exceptionally well in terms of both Precision and Recall. As shown in Table \ref{tab:comparisonProstate}, classical models outperform the quantum ones in all the cases except when using 4 features, where QSVM achieves better performance. Similar observations can also be done by comparing models in terms of F1-score, as shown in Figure \ref{fig:Prostate_classical}. However, the difference in F1 score in classical models between the training and validation sets can reach up to 20\%, and up to 10\% between the training and test sets, indicating a potential risk of overfitting.

\subsection{Discussion}

Evaluating the performance of two quantum models, QNN and QSVM, alongside several classical models --- Logistic Regression, Decision Tree, Random Forest, and SVM --- on the Heart-Failure, Diabetes and Prostate Cancer datasets, some general observation can be made regarding the potential of quantum models in healthcare classification tasks.

First of all, \textbf{QSVM performs better than QNN} on these specific types of datasets with the exception of the Heart-Failure dataset. However, the relatively poor performance of QNN in some cases can be attributed to \textbf{overfitting}, as evidenced by significant discrepancies between training and test results.

Then, for this type of dataset, classical models are frequently affected by overfitting, which reduces their efficiency in classifying new samples.

Lastly, quantum models show better performance compared to classical models on the Heart-Failure dataset, competitive performance on the Diabetes dataset, and weaker performance on the Prostate Cancer dataset. It is possible to observe that the Heart-Failure dataset exhibits the \textbf{highest imbalance} among the three, while Prostate Cancer is the most balanced. Therefore, it can be concluded that consistent with expectations, \textbf{quantum models become increasingly advantageous as dataset imbalance increases}, effectively addressing the limitations of classical models in handling such classification tasks.

\section{Conclusions}\label{sec:conclusions}

This study evaluates the potential of quantum models, specifically QNN and QSVM, in healthcare classification tasks compared to classical models like Logistic Regression, Decision Tree, Random Forest, and SVM. Three datasets --- Heart-Failure, Diabetes, and Prostate Cancer --- were analyzed to evaluate model performance across varying dataset characteristics, including class imbalance.

The results show several key insights. First of all, QSVM  outperforms QNN in most configurations, demonstrating robustness in handling imbalanced datasets and achieving high Recall. Secondly, while QNN models show potential in catching complex patterns, their performance is limited by overfitting, as evidenced by discrepancies between validation and test metrics. Enhancing the hyperparameter optimization process, e.g., exploring alternative ansatz designs, and improving the strategy for selecting the best model may mitigate. 
Finally, quantum models exhibit better performance compared to classical models on the highly imbalanced Heart-Failure dataset, where classical models presents limitations, especially in terms of Recall. However, classical models provide satisfactory results on more balanced datasets like Prostate Cancer, suggesting that the advantage of quantum models grows with dataset imbalance.

Even though preliminary, these results highlight the impact of dataset characteristics in determining model performance and prove the potential of quantum models to address challenges in healthcare classification tasks, particularly in scenarios involving imbalanced datasets, leading the way for further research in this domain. 

\bibliographystyle{IEEEtran}
\bibliography{biblio}


\end{document}